\documentclass{article}

\usepackage{PRIMEarxiv}

\usepackage[utf8]{inputenc} 
\usepackage[T1]{fontenc}    
\usepackage{hyperref}       
\usepackage{url}            
\usepackage{booktabs}       
\usepackage{amsfonts}       
\usepackage{nicefrac}       
\usepackage{microtype}      
\usepackage{lipsum}
\usepackage{fancyhdr}       
\usepackage{graphicx}       
\graphicspath{{media/}}     

\title{Building artificial neural circuits for domain-general cognition: \\ a primer on brain-inspired systems-level architecture
\thanks{This manuscript is part of the AAAI 2023 Spring Symposium on the Evaluation and Design of Generalist Systems (EDGeS)}
}

\author{
  Jascha Achterberg \\
  University of Cambridge \& Intel Labs \\
  \texttt{jascha.achterberg@mrc-cbu.cam.ac.uk} \\
   \And
  Danyal Akarca \\
  University of Cambridge \\
  \texttt{danyal.akarca@mrc-cbu.cam.ac.uk} \\
   \And
  Moataz Assem \\
  University of Cambridge \\
  \texttt{moataz.assem@mrc-cbu.cam.ac.uk} \\
   \And
  Moritz Heimbach \\
  Julius-Maximilians-Universität Würzburg \\
  \texttt{moritz.heimbach@uni-wuerzburg.de} \\
   \And
  Duncan E. Astle \\
  University of Cambridge \\
  \texttt{duncan.astle@mrc-cbu.cam.ac.uk} \\
  \And
  John Duncan \\
  University of Cambridge \\
  \texttt{john.duncan@mrc-cbu.cam.ac.uk} \\
}

\begin{document}
\maketitle

\begin{abstract}
There is a concerted effort to build domain-general artificial intelligence in the form of universal neural network models with sufficient computational flexibility to solve a wide variety of cognitive tasks but without requiring fine-tuning on individual problem spaces and domains. To do this, models need appropriate priors and inductive biases, such that trained models can generalise to out-of-distribution examples and new problem sets. Here we provide an overview of the hallmarks endowing biological neural networks with the functionality needed for flexible cognition, in order to establish which features might also be important to achieve similar functionality in artificial systems. We specifically discuss the role of system-level distribution of network communication and recurrence, in addition to the role of short-term topological changes for efficient local computation. As machine learning models become more complex, these principles may provide valuable directions in an otherwise vast space of possible architectures. In addition, testing these inductive biases within artificial systems may help us to understand the biological principles underlying domain-general cognition.
\end{abstract}


\section{Introduction}

An aspiration of machine learning research is not just to create architectures capable of achieving increasingly high levels of task-specific performance, but the genesis of models able to achieve good performance across different domains simultaneously. Recent striking advances in network models have enabled them to solve many problems within a domain with just one architecture \cite{brown_language_2020,webb_emergent_2022,srivastava_beyond_2022}. Additionally, networks are increasingly acquiring multimodal capabilities \cite{xu_multimodal_2022, akkus_multimodal_2023} and learn in open-ended task environments \cite{fan_minedojo_2022,adaptive_agent_team_human-timescale_2023}. These advances provide necessary building blocks for models capable of domain general cognition, as observed in intelligent human behaviour. Crucially, these new models may be able to go beyond simple generalisation to unseen data \cite{hardt_patterns_2022}; they may be able to learn new abilities and directly abstract them, allowing for generalisation across entire input modalities and the reuse of skills learned in one domain to support learning in entirely new domains. Indeed, this parallels how children learn over the course of their own development \cite{kievit_sensitive_2020}. However, the extent to which current models can achieve this remains limited. 

For decades, neuroscientists have been focused on identifying core features of the brain’s structural and functional architecture. This allows us to connect our knowledge of human neural architectures that enable flexible domain-general cognition \cite{duncan_integrated_2020}, with ideas on how we hope to achieve similar capabilities in artificial systems. Here we provide an overview of mechanisms underlying domain-general cognition in biological neural networks to derive which features of the systems-level architecture may be important to build flexible multimodal problem-solving capabilities into artificial systems. Previously published reviews have already outlined which cognitive ideas and modules might be essential \cite{russin_deep_2020, goyal_inductive_2022, vanrullen_deep_2021, lecun_path_2022, lake_building_2017}. We aim to expand these cognitive perspectives by providing a brief introduction to the system-level network structure underlying domain-general cognition in the brain, highlighting what structural optimisation processes we think could be used in machine learning models. In this, our goal is not to hard-code brain-like anatomy into a network model’s architecture. Instead, we aim to identify computationally beneficial structural motifs which can be soft-coded into the network’s learning process to serve as helpful inductive biases or priors. As we see increasingly complex machine learning models being built as a combination of functional submodules \cite{pfeiffer_modular_2023,akkus_multimodal_2023}, we believe that the system-level priors we outline may provide helpful guidance to coordinate information flow in the most complex artificial neural networks \cite{goyal_coordination_2022}. 

\section{A core domain-general network in the brain}

The human brain, as with many complex physical systems, is economically organised to balance numerous competing objectives – including metabolic, computational, and geometric \cite{cajal_cajals_1995, bullmore_economy_2012}. These objectives have a strong influence on the topology of the brain’s network; not only is it energetically expensive to fully build and sustain neural connections \cite{raichle_appraising_2002,tomasi_energetic_2013} but it is highly costly to constantly communicate signals between neurons and assemblies of neurons, particularly over longer distances \cite{levy_communication_2021}. Owing to its size, complexity, and these economic considerations, it is infeasible for each neural region to communicate directly with every other region equivalently \cite{horvat_spatial_2016}. To avoid this problem, evolutionary pressures have guided the brain towards a modularised network, with modules of very strong local connectivity and high-connection hub nodes connecting across these modules \cite{suarez_connectomics-based_2022,luppi_synergistic_2022}. Networks with this structure are described as having “small-world” characteristics, defined as having concurrently a highly clustered topology and short path lengths, meeting a balance between totally random versus regular networks \cite{bassett_small-world_2006,bassett_small-world_2017}. Small-world structures are commonly found in distributed systems under resource constraints, showing patterns of locally specialised computation alongside good propagation of signals within and between hubs. In brains, this locality of computations results in concentration of specific cognitive function within specific anatomical regions. Specialised regions act as foci for cognitive functions like sensory processing, semantic knowledge, and language abilities \cite{kaas_organization_2001, ralph_neural_2017, skeide_ontogeny_2016}. These are likely semi-specialised, meaning that they mostly focus on unique local computation but also partially integrate meaningful information across areas and domains \cite{atilgan_integration_2018,steinmetz_distributed_2019}.

This picture of a functionally modular system becomes more nuanced when we consider human domain-general cognition. As the tasks to be solved become more complicated, the brain increasingly abandons solely relying on its specialised modular structure. Instead, neural architectures must increasingly integrate signals across modules \cite{power_evidence_2013} and rely on its Multiple Demand system (MD) \cite{duncan_integrated_2020, assem_domain-general_2020}. This is a core network in the brain (depicted in Figure 1A) which is highly active when a complex task of any nature is solved. It is thought that the MD system serves as a central processing unit, receiving information from more specialised input notes, to compress it into meaningful abstract representations on which it can run problem solving algorithms (schematic shown in Figure 1B). It also plays a central role in controlling information in other brain regions, using knowledge and complex analysis of the situation to control thought processes in specialised brain regions through top-down control processes \cite{duncan_integrated_2020, deco_revisiting_2021,dehaene_neuronal_1998, miller_integrative_2001,norman_attention_1986}. Ultimately it is this central processing circuit that likely gives the primate brain the ability to have abstract thoughts used to solve complex problems to reach long-term goals.

\begin{figure*}
  \centering
  \includegraphics[width=15cm]{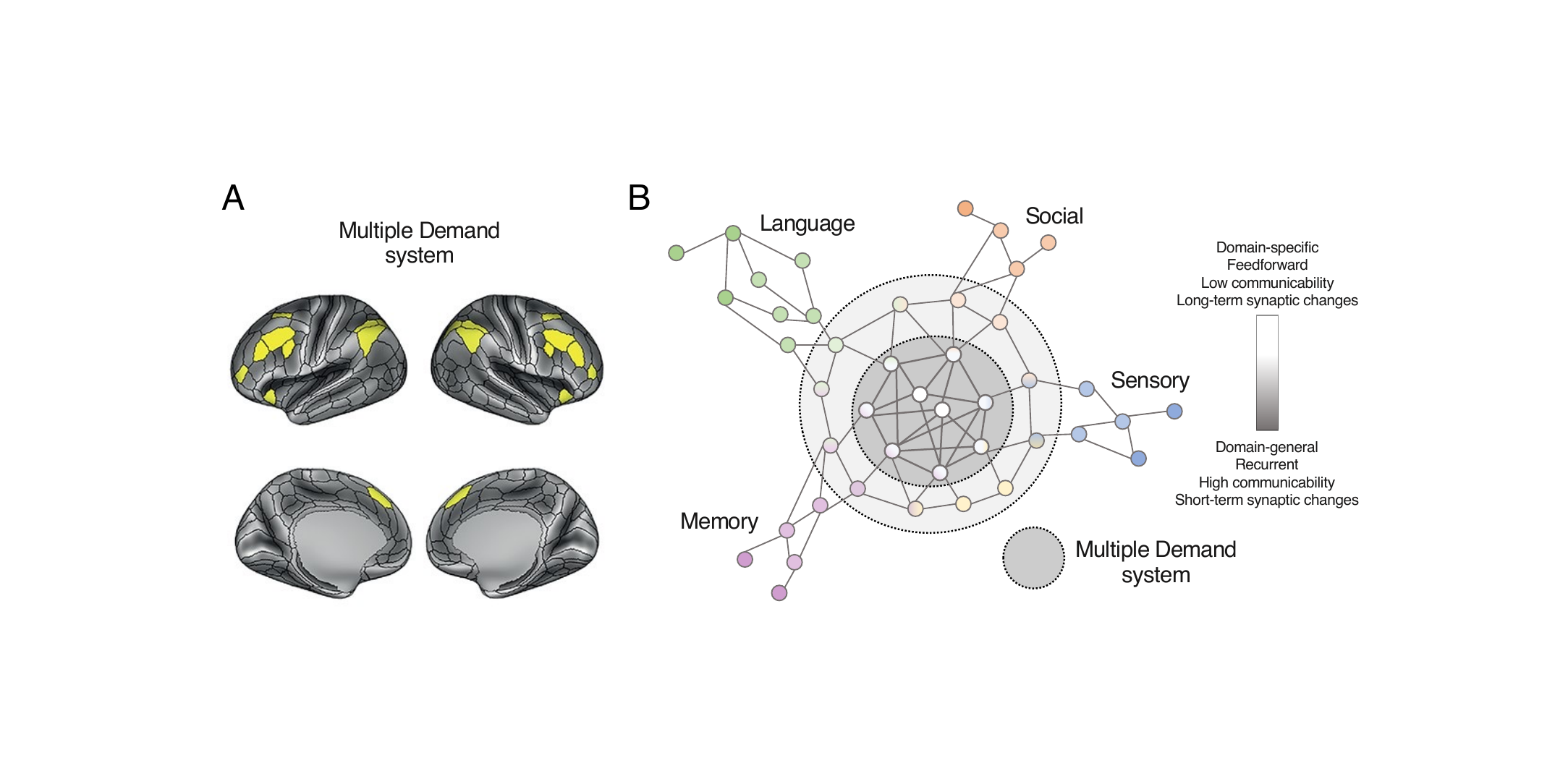}
  \vspace{-40pt}
  \caption{A - The cortical areas forming the core Multiple Demand system in the human brain, from \cite{assem_domain-general_2020}. B – Schematic depiction of a systems-level view of the brain. The Multiple Demand system lies at the core of information processing in the brain, exchanging inputs with more specialised regions such as language, memory, sensory and social processing. Due to its central position, the MD core can influence computations in multiple specialised areas by broadcasting information it constructed from integrating across domains back to specialised regions, e.g., influencing perception by abstract understanding of the environment / situation at large. Refer to \cite{assem_domain-general_2020} for detailed anatomical perspective of the MD system's core and penumbra regions not discussed here.}
\end{figure*}

What are the key principles underpinning this system? In the following we will discuss three computational / structural motifs which vary across the hierarchy from specialised regions to the integrative MD system, which allow the network to show domain-general cognitive skills. These are: Recurrence, communicability, and short-term topological changes. We will review each of these in terms of their relevance in biological networks before then discussing possible directions for artificial implementations, in the context of related existing implementations.

\section{Computational motifs supporting domain-general cognition}
\subsection{Global recurrence}

Computations in functionally more specialised regions depend strongly on a feed-forward structure that extracts increasingly abstract features from sensory inputs \cite{grill-spector_human_2004, hackett_information_2011, mashour_conscious_2020}. Much work shows how this process can be modelled using an artificial feed-forward network \cite{schrimpf_brain-score_2018, lindsay_convolutional_2021}. While there is also recurrent processing in these specialised systems \cite{grill-spector_human_2004, hackett_information_2011, kietzmann_recurrence_2019}, the recurrent loops in these systems are likely very local and cover relatively short distances and timescales. This means that a signal sent from a node will only travel a short path before arriving back at its starting point. As we move towards more integrative and domain-general cognition, recurrent connections become a hallmark feature of the brain’s systems-level design. The frontal cortex, where a large part of the MD system lies, is often thought of as implementing recurrent loops for abstract information processing \cite{mashour_conscious_2020, miller_rules_2007}. Importantly, these loops not only process information locally but also broadcast information widely across the brain, influencing and controlling computations in specialised regions. It does so by not only having local recurrent connections within the circuit but also many loops spanning large distances in the brain, reaching out to nodes which lie far outside the core \cite{miller_integrative_2001,munakata_unified_2011,mashour_conscious_2020,dehaene_neuronal_1998}. With nodes widely distributed over the cortex, coupled to strong communication between these nodes, the MD system is well positioned for widespread integration and communication. A large set of recursive processing loops with varied scales in terms of time and spatial distance likely facilitate the MD system’s abstract domain-general processing and deliver the ability to coordinate computation in a large distributed system \cite{duncan_integrated_2020}.

The use of recurrent loops in artificial neural networks has a long history \cite{schmidhuber_annotated_2022}. They proved to be useful tools for processing and predicting time series data but also suffered from problems of vanishing gradient and computational complexity when capturing long-range dependencies in the input \cite{hochreiter_long_1997,vaswani_attention_2017,fawaz_deep_2019}. To avoid these issues, feed-forward based architectures can be used as substitutes \cite{fawaz_deep_2019} and various attention-based architectures have recently been very effective in capturing dependencies in language time courses and multiple other modalities \cite{vaswani_attention_2017, tay_efficient_2022}. This works by inputting an entire time series in a single time step so that the attention mechanism learns the relationship between timesteps without needing to hold past time points in memory. While these architectures likely can be good substitutes for the local recurrent loops, we believe that ultimately, researchers are going to have to find a way to also introduce global recurrent loops to arrive at domain-general cognition in artificial systems. Approaches like weight-sharing in deep models paired with skip-connections may allow us to mimic a recurrent process in a regular forward pass but it seems likely that alternative ways will be needed to allow abstract multimodal knowledge to be broadcasted through the network to inform distributed computations. This seems even more timely now that models generate impressive responses to inputs such as images or language \cite{brown_language_2020,rombach_high-resolution_2022}, but struggle to be constrained by meaningful world models (e.g., intuitive physics, \cite{lake_building_2017}). Instead, researchers rely on human feedback signals in the training pipeline \cite{ouyang_training_2022}. As such, machine learning models may need to be adapted to allow for the introduction of a global recurrent architecture similar to the MD system.

\subsection{Communicability in large scale networks}

For any complex network which is concerned with processing information, it is of central importance to optimise how signals are communicated between the nodes within the network \cite{estrada_physics_2012}. This becomes an increasingly challenging problem as a network grows, leaving nodes to only be able to communicate with a smaller proportion of the network. This limited communication capacity naturally leads to variation in terms of how much information is exchanged between different pairs of nodes across the network. This results in a very real challenge for any large-scale network system to optimise its structure to integrate information most effectively and efficiently across its functional hubs. This is constrained, ultimately, by the topological arrangement of the network. The idea of how much information is exchanged between nodes is captured by the concept of communicability \cite{estrada_physics_2012,crofts_weighted_2009,srivastava_models_2020} and is a highly effective framework to understand how the structure of the brain guides function \cite{goni_resting-brain_2014,seguin_inferring_2019,betzel_multi-policy_2022, griffa_evolution_2022, avena-koenigsberger_communication_2018,avena-koenigsberger_spectrum_2019, laughlin_communication_2003}. Specifically, across the brain’s complex network, regions vary in terms of how well they can communicate to other regions, and the macro-scale dynamics and capabilities of the brain will be determined by this interareal communication. This heterogeneous communicability becomes especially interesting when one considers how system-level communication link to domain-general cognition. In the previous section, we described how more specialised regions tend to have a mostly feed-forward structure with some local recurrence. As such, information tends to be communicated locally between adjacent and functionally related regions. This changes as information approaches the domain-general MD system with its wider communicative influence. In its central position, the MD system not only receives information from all over the brain but utilises its widespread connectivity as global recurrent loops to broadcast processed information to a distributed set of brain regions \cite{mashour_conscious_2020,duncan_integrated_2020,dehaene_neuronal_1998}. On the systems-level perspective of the brain, a given region’s communicative structure heavily depends on its functional role and hence its degree of specialisation. 

The concept of heterogeneous communicability between regions and modules of the brain has not been particularly relevant in artificial neural network architectures which were state-of-the-art until very recently. Take convolutional neural networks (CNNs) as an example. In CNNs, which dominated processing of visual information for several years \cite{schmidhuber_annotated_2022}, information is mostly passed along from layer to layer in a relatively even fashion. This means regions do not stick out has having a particular communicative ability (though see work like \cite{shrivastava_beyond_2017} for interesting communicative extensions of CNNs). However, this is changing with new architectures which have been growing in scale \cite{kaplan_scaling_2020}. Especially for network models which utilise multiple modalities, architectures have increasingly been created by combining existing pre-trained models into more complex modular architectures \cite{rombach_high-resolution_2022, akkus_multimodal_2023}. Once we build complex system like these, it becomes increasingly important to not only think about which models to combine, but also how to combine them. This means that the communicability between parts of the network can be optimised to achieve better information flow between components and hence improve performance. A first step in this direction was made by a multimodal transformer model which outperformed prior networks by introducing a set of special bridge layers to connect two modality specific models. These bridge layers allow the model to learn a communicative structure in which abstract semantic knowledge is gradually merged across modalities. This increased performance in several relevant benchmark tasks \cite{xu_bridgetower_2023}. In addition, other implementations have shown that bringing ideas from highly communicative small world graph structures into a Transformer’s attention mechanism can help with processing longer sequences \cite{zaheer_big_2021}. In simple recurrent neural networks, we also have seen that system-level communicability can easily be used as a regularisation term to optimise the communicative structure of a sparsely connected network to arrive at a network with many brain-like structural and functional properties \cite{achterberg_spatially-embedded_2022}. As network models grow in complexity and increasingly make use of composite structures which combine sub models into larger networks, it will be important to fine tune the communicative structure of a network. Having good priors and inductive biases for these linkages can help circumvent problems arising from adding the extensive set of connections it would require to fully connect multiple models which already have a complex structure themselves. Following this line of thinking, we believe that making use of work on communicability and how it can be optimised in complex networks will be of central importance to inform model building on a systems-level.

\subsection{Short-term topological changes}

The discussion so far has focused on how the systems-level network structure of the brain and the unique communicative structure of its MD system play a key role in the domain-general cognition we see in humans. An important element of its flexible and multimodal information processing capabilities is how the MD system’s network structure is not fully fixed but often rapidly changing. This means that while the MD system is running multimodal computations internally, the connections between its neurons are in continuous flux. As such, the general problem-solving ability of this network is assumed to be due to its inherent flexibility. In it, local computations are organised by rapid changes to the network structure \cite{stokes_activity-silent_2015,tang_prefrontal_2022,garcia-cabezas_mirror_2017}, often called short-term plasticity. This allows the network to continuously reassign its neurons and modules new computational roles while solving complex sequential problems \cite{duncan_adaptive_2001,miller_integrative_2001,crowe_rapid_2010,meyers_dynamic_2008,achterberg_one-shot_2022}. Rapid topological changes are likely induced by local learning rules which supplement the more long-term optimisation of the global network structure. These mechanisms likely underly a multitude of complex abilities of the human brain \cite{assem_domain-general_2020,duncan_multiple-demand_2010} and some of them strongly overlap with timely discussions in machine learning. As one example, research points to the fact that the MD system uses its short-term dynamics for attentional control, to focus on information which is relevant for the current operation \cite{sakagami_encoding_1994,rainer_selective_1998, buschman_synchronous_2012} and break complex pieces of information down into simple computable bits \cite{duncan_integrated_2020} – a function which has played a central role in machine learning discussion recently \cite{shrivastava_beyond_2017,lindsay_attention_2020}. Another example is the MD system’s ability to construct abstract representations of problems \cite{wallis_single_2001} to then tie observed stimuli rapidly to their roles in this abstract problem representation \cite{duncan_integrated_2020,achterberg_one-shot_2022}, a phenomenon going by the name of variable binding \cite{smolensky_tensor_1990} or meta-learning \cite{botvinick_reinforcement_2019}. These are very related to few-shot learning \cite{brown_language_2020} and in-context learning abilities \cite{von_oswald_transformers_2022} observed in large Transformer models. 

As we already see foundations of these skills emerging in currently existing architectures it is reasonable to believe that they will continue to improve purely by scaling existing architectures \cite{kaplan_scaling_2020}. In this case models would use their unit activations to implement rapid in-context learning and it has been shown that this can work well without any short-term synaptic changes \cite{wang_prefrontal_2018}. In fact, even in the brain many complex computations are likely facilitated due to dynamics of the network activations which do not necessarily have to rely on changes in the network structure \cite{vyas_computation_2020}. But once computations reach the scale of using network-wide attention processes to controlling the flow of information across the entire brain network and flexibly combining task modules to solve the task at hand \cite{duncan_integrated_2020,buschman_goal-direction_2014,macdowell_multiplexed_2023}, rapid topological network changes might be necessary for domain-general computations \cite{stokes_activity-silent_2015,duncan_integrated_2020}. Reaching this level of flexible and multimodal cognition might not be possible in current static architectures and hence might require us to allow models to modify some of their connections in the moment through local learning rules. Some work in smaller network models is highlighting how local learning mechanisms can complement network-wide optimisation processes \cite{whittington_tolman-eichenbaum_2020,dekker_determinants_2022} with relevant comparisons to Transformer implementations \cite{whittington_relating_2022}. Other examples point to how local learning rules and single neuron-based optimisation principles by themselves can be sufficient to solve meaningful cognitive tasks \cite{masse_circuit_2019,falandays_potential_2023}. In addition, we have seen how standard network optimisers can be updated with certainty judgements to support rapid relational learning \cite{nelli_neural_2023}. If we could scale these rapid learning dynamics to large Transformer models, this might allow models to flexibly combine abstract task structures with capabilities learned in the past, to flexibly apply skills across modalities in a truly domain-general way. One research direction which might support rapid learning processes is work on using local loss functions and learning mechanisms to substitute costly global optimisation processes \cite{lowe_putting_2020,ren_scaling_2023, hinton_forward-forward_2022}. Combining these local optimisation processes with more wide-spread recurrent loops and an optimised communicative structure in large networks might bring us closer to observing flexible domain-general cognition in artificial neural networks.

\section{Conclusion}

We believe that in the pursuit of building artificial intelligence which is able to engage in domain-general problem solving, a systems-level view of the human brain will provide useful guidance \cite{hassabis_neuroscience-inspired_2017,zador_toward_2023}. We believe this will become increasingly relevant as AI systems become more and more complex. The topics of recurrence, communication and rapid structural changes are particularly relevant at the current point due to their central role in theories of domain-general cognition in the brain and their links to existing works in neural network models. As such, they might be key drivers behind efficient and flexible information processing in large multimodal networks. But we do not believe that any of these features should be fully hard-coded – instead we should think of them as useful priors and inductive biases which can guide complex learning processes. Ultimately, bringing these features into machine learning models opens up the perspective of not only improving the performance of artificial neural networks but also for us to understand which core principles underly domain-general and multimodal computations in neural networks - may these be biological or artificial.

\section{Acknowledgments}
J.A., Da.A., M.A., Du.A., and J.D. are supported by UKRI MRC funding and as a result the authors have applied a Creative Commons Attribution (CC BY) license to this manuscript for the purpose of open access. J.A. receives a Gates Cambridge Scholarship. Da.A. receives a Cambridge Trust Vice Chancellor’s Scholarship. Da.A. and Du.A. are both supported by the James S. McDonnell Foundation Opportunity Award. J.A. was a research intern at Intel Labs at the time of writing this manuscript.

\bibliographystyle{unsrt}  
\bibliography{references}

\begin{thebibliography}{10}

\bibitem{brown_language_2020}
Tom~B. Brown, Benjamin Mann, Nick Ryder, Melanie Subbiah, Jared Kaplan,
  Prafulla Dhariwal, Arvind Neelakantan, Pranav Shyam, Girish Sastry, Amanda
  Askell, Sandhini Agarwal, Ariel Herbert-Voss, Gretchen Krueger, Tom Henighan,
  Rewon Child, Aditya Ramesh, Daniel~M. Ziegler, Jeffrey Wu, Clemens Winter,
  Christopher Hesse, Mark Chen, Eric Sigler, Mateusz Litwin, Scott Gray,
  Benjamin Chess, Jack Clark, Christopher Berner, Sam McCandlish, Alec Radford,
  Ilya Sutskever, and Dario Amodei.
\newblock Language {Models} are {Few}-{Shot} {Learners}, July 2020.
\newblock arXiv:2005.14165 [cs].

\bibitem{webb_emergent_2022}
Taylor Webb, Keith~J. Holyoak, and Hongjing Lu.
\newblock Emergent {Analogical} {Reasoning} in {Large} {Language} {Models},
  December 2022.
\newblock arXiv:2212.09196 [cs].

\bibitem{srivastava_beyond_2022}
Aarohi Srivastava, Abhinav Rastogi, Abhishek Rao, Abu Awal~Md Shoeb, Abubakar
  Abid, Adam Fisch, Adam~R. Brown, Adam Santoro, Aditya Gupta, Adrià
  Garriga-Alonso, Agnieszka Kluska, Aitor Lewkowycz, Akshat Agarwal, Alethea
  Power, et~al.
\newblock Beyond the {Imitation} {Game}: {Quantifying} and extrapolating the
  capabilities of language models, June 2022.
\newblock arXiv:2206.04615 [cs, stat].

\bibitem{xu_multimodal_2022}
Peng Xu, Xiatian Zhu, and David~A. Clifton.
\newblock Multimodal {Learning} with {Transformers}: {A} {Survey}, June 2022.
\newblock arXiv:2206.06488 [cs].

\bibitem{akkus_multimodal_2023}
Cem Akkus, Luyang Chu, Vladana Djakovic, Steffen Jauch-Walser, Philipp Koch,
  Giacomo Loss, Christopher Marquardt, Marco Moldovan, Nadja Sauter, Maximilian
  Schneider, Rickmer Schulte, Karol Urbanczyk, Jann Goschenhofer, Christian
  Heumann, Rasmus Hvingelby, Daniel Schalk, and Matthias Aßenmacher.
\newblock Multimodal {Deep} {Learning}, January 2023.
\newblock arXiv:2301.04856 [cs, stat].

\bibitem{fan_minedojo_2022}
Linxi Fan, Guanzhi Wang, Yunfan Jiang, Ajay Mandlekar, Yuncong Yang, Haoyi Zhu,
  Andrew Tang, De-An Huang, Yuke Zhu, and Anima Anandkumar.
\newblock {MineDojo}: {Building} {Open}-{Ended} {Embodied} {Agents} with
  {Internet}-{Scale} {Knowledge}, November 2022.
\newblock arXiv:2206.08853 [cs].

\bibitem{adaptive_agent_team_human-timescale_2023}
{Adaptive Agent Team}, Jakob Bauer, Kate Baumli, Satinder Baveja, Feryal
  Behbahani, Avishkar Bhoopchand, Nathalie Bradley-Schmieg, Michael Chang,
  Natalie Clay, Adrian Collister, Vibhavari Dasagi, Lucy Gonzalez, Karol
  Gregor, Edward Hughes, Sheleem Kashem, Maria Loks-Thompson, Hannah Openshaw,
  Jack Parker-Holder, Shreya Pathak, Nicolas Perez-Nieves, Nemanja Rakicevic,
  Tim Rocktäschel, Yannick Schroecker, Jakub Sygnowski, Karl Tuyls, Sarah
  York, Alexander Zacherl, and Lei Zhang.
\newblock Human-{Timescale} {Adaptation} in an {Open}-{Ended} {Task} {Space},
  January 2023.
\newblock arXiv:2301.07608 [cs].

\bibitem{hardt_patterns_2022}
Moritz Hardt and Benjamin Recht.
\newblock {\em Patterns, predictions, and actions: a story about machine
  learning}.
\newblock Princeton University Press, Princeton, 2022.

\bibitem{kievit_sensitive_2020}
Rogier~A Kievit.
\newblock Sensitive periods in cognitive development: a mutualistic
  perspective.
\newblock {\em Current Opinion in Behavioral Sciences}, 36:144--149, December
  2020.

\bibitem{duncan_integrated_2020}
John Duncan, Moataz Assem, and Sneha Shashidhara.
\newblock Integrated {Intelligence} from {Distributed} {Brain} {Activity}.
\newblock {\em Trends in Cognitive Sciences}, 24(10):838--852, October 2020.

\bibitem{russin_deep_2020}
Jacob Russin, Randall~C O’Reilly, and Yoshua Bengio.
\newblock {DEEP} {LEARNING} {NEEDS} {A} {PREFRONTAL} {CORTEX}.
\newblock {\em “Bridging AI and Cognitive Science” (ICLR 2020)}, 2020.

\bibitem{goyal_inductive_2022}
Anirudh Goyal and Yoshua Bengio.
\newblock Inductive biases for deep learning of higher-level cognition.
\newblock {\em Proceedings of the Royal Society A: Mathematical, Physical and
  Engineering Sciences}, 478(2266):20210068, October 2022.
\newblock Publisher: Royal Society.

\bibitem{vanrullen_deep_2021}
Rufin VanRullen and Ryota Kanai.
\newblock Deep learning and the {Global} {Workspace} {Theory}.
\newblock {\em Trends in Neurosciences}, 44(9):692--704, September 2021.

\bibitem{lecun_path_2022}
Yann LeCun.
\newblock A {Path} {Towards} {Autonomous} {Machine} {Intelligence} {Version}
  0.9.2, 2022-06-27, June 2022.

\bibitem{lake_building_2017}
Brenden~M. Lake, Tomer~D. Ullman, Joshua~B. Tenenbaum, and Samuel~J. Gershman.
\newblock Building machines that learn and think like people.
\newblock {\em Behavioral and Brain Sciences}, 40:e253, 2017.
\newblock Publisher: Cambridge University Press.

\bibitem{pfeiffer_modular_2023}
Jonas Pfeiffer, Sebastian Ruder, Ivan Vulić, and Edoardo~Maria Ponti.
\newblock Modular {Deep} {Learning}, February 2023.
\newblock arXiv:2302.11529 [cs].

\bibitem{goyal_coordination_2022}
Anirudh Goyal, Aniket Didolkar, Alex Lamb, Kartikeya Badola, Nan~Rosemary Ke,
  Nasim Rahaman, Jonathan Binas, Charles Blundell, Michael Mozer, and Yoshua
  Bengio.
\newblock Coordination {Among} {Neural} {Modules} {Through} a {Shared} {Global}
  {Workspace}, March 2022.
\newblock arXiv:2103.01197 [cs, stat].

\bibitem{cajal_cajals_1995}
Santiago Ramon~y Cajal, Neely Swanson, Larry~W. Swanson, Santiago Ramon~y
  Cajal, Neely Swanson, and Larry~W. Swanson.
\newblock {\em Cajal's {Histology} of the {Nervous} {System} of {Man} and
  {Vertebrates}}.
\newblock History of {Neuroscience}. Oxford University Press, Oxford, New York,
  April 1995.

\bibitem{bullmore_economy_2012}
Ed~Bullmore and Olaf Sporns.
\newblock The economy of brain network organization.
\newblock {\em Nature Reviews Neuroscience}, 13(5):336--349, May 2012.
\newblock Number: 5 Publisher: Nature Publishing Group.

\bibitem{raichle_appraising_2002}
Marcus~E. Raichle and Debra~A. Gusnard.
\newblock Appraising the brain's energy budget.
\newblock {\em Proceedings of the National Academy of Sciences},
  99(16):10237--10239, August 2002.
\newblock Publisher: Proceedings of the National Academy of Sciences.

\bibitem{tomasi_energetic_2013}
Dardo Tomasi, Gene-Jack Wang, and Nora~D. Volkow.
\newblock Energetic cost of brain functional connectivity.
\newblock {\em Proceedings of the National Academy of Sciences},
  110(33):13642--13647, August 2013.
\newblock Publisher: Proceedings of the National Academy of Sciences.

\bibitem{levy_communication_2021}
William~B Levy and Victoria~G. Calvert.
\newblock Communication consumes 35 times more energy than computation in the
  human cortex, but both costs are needed to predict synapse number.
\newblock {\em Proceedings of the National Academy of Sciences},
  118(18):e2008173118, May 2021.
\newblock Publisher: Proceedings of the National Academy of Sciences.

\bibitem{horvat_spatial_2016}
Szabolcs Horvát, Răzvan Gămănuț, Mária Ercsey-Ravasz, Loïc Magrou,
  Bianca Gămănuț, David C.~Van Essen, Andreas Burkhalter, Kenneth Knoblauch,
  Zoltán Toroczkai, and Henry Kennedy.
\newblock Spatial {Embedding} and {Wiring} {Cost} {Constrain} the {Functional}
  {Layout} of the {Cortical} {Network} of {Rodents} and {Primates}.
\newblock {\em PLOS Biology}, 14(7):e1002512, July 2016.
\newblock Publisher: Public Library of Science.

\bibitem{suarez_connectomics-based_2022}
Laura~E Suarez, Yossi Yovel, Martijn~P van~den Heuvel, Olaf Sporns, Yaniv
  Assaf, Guillaume Lajoie, and Bratislav Misic.
\newblock A connectomics-based taxonomy of mammals.
\newblock {\em eLife}, 11:e78635, November 2022.
\newblock Publisher: eLife Sciences Publications, Ltd.

\bibitem{luppi_synergistic_2022}
Andrea~I. Luppi, Pedro A.~M. Mediano, Fernando~E. Rosas, Negin Holland, Tim~D.
  Fryer, John~T. O’Brien, James~B. Rowe, David~K. Menon, Daniel Bor, and
  Emmanuel~A. Stamatakis.
\newblock A synergistic core for human brain evolution and cognition.
\newblock {\em Nature Neuroscience}, 25(6):771--782, June 2022.
\newblock Number: 6 Publisher: Nature Publishing Group.

\bibitem{bassett_small-world_2006}
Danielle~Smith Bassett and Ed~Bullmore.
\newblock Small-{World} {Brain} {Networks}.
\newblock {\em The Neuroscientist}, 12(6):512--523, December 2006.
\newblock Publisher: SAGE Publications Inc STM.

\bibitem{bassett_small-world_2017}
Danielle~S. Bassett and Edward~T. Bullmore.
\newblock Small-{World} {Brain} {Networks} {Revisited}.
\newblock {\em The Neuroscientist}, 23(5):499--516, October 2017.
\newblock Publisher: SAGE Publications Inc STM.

\bibitem{kaas_organization_2001}
Jon~H Kaas and Christine~E Collins.
\newblock The organization of sensory cortex.
\newblock {\em Current Opinion in Neurobiology}, 11(4):498--504, August 2001.

\bibitem{ralph_neural_2017}
Matthew A.~Lambon Ralph, Elizabeth Jefferies, Karalyn Patterson, and Timothy~T.
  Rogers.
\newblock The neural and computational bases of semantic cognition.
\newblock {\em Nature Reviews Neuroscience}, 18(1):42--55, January 2017.
\newblock Number: 1 Publisher: Nature Publishing Group.

\bibitem{skeide_ontogeny_2016}
Michael~A. Skeide and Angela~D. Friederici.
\newblock The ontogeny of the cortical language network.
\newblock {\em Nature Reviews Neuroscience}, 17(5):323--332, May 2016.
\newblock Number: 5 Publisher: Nature Publishing Group.

\bibitem{atilgan_integration_2018}
Huriye Atilgan, Stephen~M. Town, Katherine~C. Wood, Gareth~P. Jones, Ross~K.
  Maddox, Adrian K.~C. Lee, and Jennifer~K. Bizley.
\newblock Integration of {Visual} {Information} in {Auditory} {Cortex}
  {Promotes} {Auditory} {Scene} {Analysis} through {Multisensory} {Binding}.
\newblock {\em Neuron}, 97(3):640--655.e4, February 2018.

\bibitem{steinmetz_distributed_2019}
Nicholas~A. Steinmetz, Peter Zatka-Haas, Matteo Carandini, and Kenneth~D.
  Harris.
\newblock Distributed coding of choice, action and engagement across the mouse
  brain.
\newblock {\em Nature}, 576(7786):266--273, December 2019.
\newblock Number: 7786 Publisher: Nature Publishing Group.

\bibitem{power_evidence_2013}
Jonathan~D. Power, Bradley~L. Schlaggar, Christina~N. Lessov-Schlaggar, and
  Steven~E. Petersen.
\newblock Evidence for {Hubs} in {Human} {Functional} {Brain} {Networks}.
\newblock {\em Neuron}, 79(4):798--813, August 2013.

\bibitem{assem_domain-general_2020}
Moataz Assem, Matthew~F Glasser, David~C Van~Essen, and John Duncan.
\newblock A {Domain}-{General} {Cognitive} {Core} {Defined} in {Multimodally}
  {Parcellated} {Human} {Cortex}.
\newblock {\em Cerebral Cortex}, 30(8):4361--4380, June 2020.

\bibitem{deco_revisiting_2021}
Gustavo Deco, Diego Vidaurre, and Morten~L. Kringelbach.
\newblock Revisiting the global workspace orchestrating the hierarchical
  organization of the human brain.
\newblock {\em Nature Human Behaviour}, 5(4):497--511, April 2021.
\newblock Number: 4 Publisher: Nature Publishing Group.

\bibitem{dehaene_neuronal_1998}
Stanislas Dehaene, Michel Kerszberg, and Jean-Pierre Changeux.
\newblock A neuronal model of a global workspace in effortful cognitive tasks.
\newblock {\em Proceedings of the National Academy of Sciences},
  95(24):14529--14534, November 1998.
\newblock Publisher: Proceedings of the National Academy of Sciences.

\bibitem{miller_integrative_2001}
Earl~K. Miller and Jonathan~D. Cohen.
\newblock An {Integrative} {Theory} of {Prefrontal} {Cortex} {Function}.
\newblock {\em Annual Review of Neuroscience}, 24(1):167--202, 2001.
\newblock \_eprint: https://doi.org/10.1146/annurev.neuro.24.1.167.

\bibitem{norman_attention_1986}
Donald~A. Norman and Tim Shallice.
\newblock Attention to {Action}.
\newblock In Richard~J. Davidson, Gary~E. Schwartz, and David Shapiro, editors,
  {\em Consciousness and {Self}-{Regulation}: {Advances} in {Research} and
  {Theory} {Volume} 4}, pages 1--18. Springer US, Boston, MA, 1986.

\bibitem{grill-spector_human_2004}
Kalanit Grill-Spector and Rafael Malach.
\newblock The {Human} {Visual} {Cortex}.
\newblock {\em Annual Review of Neuroscience}, 27(1):649--677, 2004.
\newblock \_eprint: https://doi.org/10.1146/annurev.neuro.27.070203.144220.

\bibitem{hackett_information_2011}
Troy~A. Hackett.
\newblock Information flow in the auditory cortical network.
\newblock {\em Hearing Research}, 271(1):133--146, January 2011.

\bibitem{mashour_conscious_2020}
George~A. Mashour, Pieter Roelfsema, Jean-Pierre Changeux, and Stanislas
  Dehaene.
\newblock Conscious {Processing} and the {Global} {Neuronal} {Workspace}
  {Hypothesis}.
\newblock {\em Neuron}, 105(5):776--798, March 2020.

\bibitem{schrimpf_brain-score_2018}
Martin Schrimpf, Jonas Kubilius, Ha~Hong, Najib~J. Majaj, Rishi Rajalingham,
  Elias~B. Issa, Kohitij Kar, Pouya Bashivan, Jonathan Prescott-Roy, Kailyn
  Schmidt, Daniel L.~K. Yamins, and James~J. DiCarlo.
\newblock Brain-{Score}: {Which} {Artificial} {Neural} {Network} for {Object}
  {Recognition} is most {Brain}-{Like}?, September 2018.
\newblock bioRxiv: 407007 Section: New Results.

\bibitem{lindsay_convolutional_2021}
Grace~W. Lindsay.
\newblock Convolutional {Neural} {Networks} as a {Model} of the {Visual}
  {System}: {Past}, {Present}, and {Future}.
\newblock {\em Journal of Cognitive Neuroscience}, 33(10):2017--2031, September
  2021.

\bibitem{kietzmann_recurrence_2019}
Tim~C. Kietzmann, Courtney~J. Spoerer, Lynn K.~A. Sörensen, Radoslaw~M. Cichy,
  Olaf Hauk, and Nikolaus Kriegeskorte.
\newblock Recurrence is required to capture the representational dynamics of
  the human visual system.
\newblock {\em Proceedings of the National Academy of Sciences},
  116(43):21854--21863, October 2019.
\newblock Publisher: Proceedings of the National Academy of Sciences.

\bibitem{miller_rules_2007}
Earl~K. Miller and Timothy~J. Buschman.
\newblock Rules through {Recursion}: {How} {Interactions} between the {Frontal}
  {Cortex} and {Basal} {Ganglia} {May} {Build} {Abstract}, {Complex} {Rules}
  from {Concrete}, {Simple} {Ones}.
\newblock In Silvia~A. Bunge and Jonathan~D. Wallis, editors, {\em Neuroscience
  of {Rule}-{Guided} {Behavior}}, page~0. Oxford University Press, November
  2007.

\bibitem{munakata_unified_2011}
Yuko Munakata, Seth~A. Herd, Christopher~H. Chatham, Brendan~E. Depue, Marie~T.
  Banich, and Randall~C. O’Reilly.
\newblock A unified framework for inhibitory control.
\newblock {\em Trends in Cognitive Sciences}, 15(10):453--459, October 2011.

\bibitem{schmidhuber_annotated_2022}
Juergen Schmidhuber.
\newblock Annotated {History} of {Modern} {AI} and {Deep} {Learning}, December
  2022.
\newblock arXiv:2212.11279 [cs].

\bibitem{hochreiter_long_1997}
Sepp Hochreiter and Jürgen Schmidhuber.
\newblock Long {Short}-{Term} {Memory}.
\newblock {\em Neural Computation}, 9(8):1735--1780, November 1997.
\newblock Conference Name: Neural Computation.

\bibitem{vaswani_attention_2017}
Ashish Vaswani, Noam Shazeer, Niki Parmar, Jakob Uszkoreit, Llion Jones,
  Aidan~N Gomez, Lukasz Kaiser, and Illia Polosukhin.
\newblock Attention is {All} you {Need}.
\newblock In {\em Advances in {Neural} {Information} {Processing} {Systems}},
  volume~30. Curran Associates, Inc., 2017.

\bibitem{fawaz_deep_2019}
Hassan~Ismail Fawaz, Germain Forestier, Jonathan Weber, Lhassane Idoumghar, and
  Pierre-Alain Muller.
\newblock Deep learning for time series classification: a review.
\newblock {\em Data Mining and Knowledge Discovery}, 33(4):917--963, July 2019.
\newblock arXiv:1809.04356 [cs, stat].

\bibitem{tay_efficient_2022}
Yi~Tay, Mostafa Dehghani, Dara Bahri, and Donald Metzler.
\newblock Efficient {Transformers}: {A} {Survey}.
\newblock {\em ACM Computing Surveys}, 55(6):109:1--109:28, December 2022.

\bibitem{rombach_high-resolution_2022}
Robin Rombach, Andreas Blattmann, Dominik Lorenz, Patrick Esser, and Björn
  Ommer.
\newblock High-{Resolution} {Image} {Synthesis} with {Latent} {Diffusion}
  {Models}, April 2022.
\newblock arXiv:2112.10752 [cs].

\bibitem{ouyang_training_2022}
Long Ouyang, Jeff Wu, Xu~Jiang, Diogo Almeida, Carroll~L. Wainwright, Pamela
  Mishkin, Chong Zhang, Sandhini Agarwal, Katarina Slama, Alex Ray, John
  Schulman, Jacob Hilton, Fraser Kelton, Luke Miller, Maddie Simens, Amanda
  Askell, Peter Welinder, Paul Christiano, Jan Leike, and Ryan Lowe.
\newblock Training language models to follow instructions with human feedback,
  March 2022.
\newblock arXiv:2203.02155 [cs].

\bibitem{estrada_physics_2012}
Ernesto Estrada, Naomichi Hatano, and Michele Benzi.
\newblock The physics of communicability in complex networks.
\newblock {\em Physics Reports}, 514(3):89--119, May 2012.

\bibitem{crofts_weighted_2009}
Jonathan~J Crofts and Desmond~J Higham.
\newblock A weighted communicability measure applied to complex brain networks.
\newblock {\em Journal of The Royal Society Interface}, 6(33):411--414, January
  2009.
\newblock Publisher: Royal Society.

\bibitem{srivastava_models_2020}
Pragya Srivastava, Erfan Nozari, Jason~Z. Kim, Harang Ju, Dale Zhou, Cassiano
  Becker, Fabio Pasqualetti, George~J. Pappas, and Danielle~S. Bassett.
\newblock Models of communication and control for brain networks: distinctions,
  convergence, and future outlook.
\newblock {\em Network Neuroscience}, 4(4):1122--1159, November 2020.

\bibitem{goni_resting-brain_2014}
Joaquín Goñi, Martijn~P. van~den Heuvel, Andrea Avena-Koenigsberger, Nieves
  Velez~de Mendizabal, Richard~F. Betzel, Alessandra Griffa, Patric Hagmann,
  Bernat Corominas-Murtra, Jean-Philippe Thiran, and Olaf Sporns.
\newblock Resting-brain functional connectivity predicted by analytic measures
  of network communication.
\newblock {\em Proceedings of the National Academy of Sciences},
  111(2):833--838, January 2014.
\newblock Publisher: Proceedings of the National Academy of Sciences.

\bibitem{seguin_inferring_2019}
Caio Seguin, Adeel Razi, and Andrew Zalesky.
\newblock Inferring neural signalling directionality from undirected structural
  connectomes.
\newblock {\em Nature Communications}, 10(1):4289, September 2019.
\newblock Number: 1 Publisher: Nature Publishing Group.

\bibitem{betzel_multi-policy_2022}
Richard~F. Betzel, Joshua Faskowitz, Bratislav Mišić, Olaf Sporns, and Caio
  Seguin.
\newblock Multi-policy models of interregional communication in the human
  connectome, May 2022.
\newblock bioRxiv: 2022.05.08.490752 Section: New Results.

\bibitem{griffa_evolution_2022}
Alessandra Griffa, Mathieu Mach, Julien Dedelley, Daniel Gutierrez-Barragan,
  Alessandro Gozzi, Gilles Allali, Joanes Grandjean, Dimitri Van~De Ville, and
  Enrico Amico.
\newblock The evolution of information transmission in mammalian brain
  networks, May 2022.
\newblock bioRxiv: 2022.05.09.491115 Section: New Results.

\bibitem{avena-koenigsberger_communication_2018}
Andrea Avena-Koenigsberger, Bratislav Misic, and Olaf Sporns.
\newblock Communication dynamics in complex brain networks.
\newblock {\em Nature Reviews Neuroscience}, 19(1):17--33, January 2018.
\newblock Number: 1 Publisher: Nature Publishing Group.

\bibitem{avena-koenigsberger_spectrum_2019}
Andrea Avena-Koenigsberger, Xiaoran Yan, Artemy Kolchinsky, Martijn P. van~den
  Heuvel, Patric Hagmann, and Olaf Sporns.
\newblock A spectrum of routing strategies for brain networks.
\newblock {\em PLOS Computational Biology}, 15(3):e1006833, March 2019.
\newblock Publisher: Public Library of Science.

\bibitem{laughlin_communication_2003}
Simon~B. Laughlin and Terrence~J. Sejnowski.
\newblock Communication in {Neuronal} {Networks}.
\newblock {\em Science}, 301(5641):1870--1874, September 2003.
\newblock Publisher: American Association for the Advancement of Science.

\bibitem{shrivastava_beyond_2017}
Abhinav Shrivastava, Rahul Sukthankar, Jitendra Malik, and Abhinav Gupta.
\newblock Beyond {Skip} {Connections}: {Top}-{Down} {Modulation} for {Object}
  {Detection}, September 2017.
\newblock arXiv:1612.06851 [cs].

\bibitem{kaplan_scaling_2020}
Jared Kaplan, Sam McCandlish, Tom Henighan, Tom~B. Brown, Benjamin Chess, Rewon
  Child, Scott Gray, Alec Radford, Jeffrey Wu, and Dario Amodei.
\newblock Scaling {Laws} for {Neural} {Language} {Models}, January 2020.
\newblock arXiv:2001.08361 [cs, stat].

\bibitem{xu_bridgetower_2023}
Xiao Xu, Chenfei Wu, Shachar Rosenman, Vasudev Lal, Wanxiang Che, and Nan Duan.
\newblock {BridgeTower}: {Building} {Bridges} {Between} {Encoders} in
  {Vision}-{Language} {Representation} {Learning}, February 2023.
\newblock arXiv:2206.08657 [cs].

\bibitem{zaheer_big_2021}
Manzil Zaheer, Guru Guruganesh, Avinava Dubey, Joshua Ainslie, Chris Alberti,
  Santiago Ontanon, Philip Pham, Anirudh Ravula, Qifan Wang, Li~Yang, and Amr
  Ahmed.
\newblock Big {Bird}: {Transformers} for {Longer} {Sequences}, January 2021.
\newblock arXiv:2007.14062 [cs, stat].

\bibitem{achterberg_spatially-embedded_2022}
Jascha Achterberg, Danyal Akarca, D.~J. Strouse, John Duncan, and Duncan~E.
  Astle.
\newblock Spatially-embedded recurrent neural networks reveal widespread links
  between structural and functional neuroscience findings, November 2022.
\newblock bioRxiv: 2022.11.17.516914 Section: New Results.

\bibitem{stokes_activity-silent_2015}
Mark~G. Stokes.
\newblock ‘{Activity}-silent’ working memory in prefrontal cortex: a
  dynamic coding framework.
\newblock {\em Trends in Cognitive Sciences}, 19(7):394--405, July 2015.

\bibitem{tang_prefrontal_2022}
Hua Tang, Mitchell~R. Riley, Balbir Singh, Xue-Lian Qi, David~T. Blake, and
  Christos Constantinidis.
\newblock Prefrontal cortical plasticity during learning of cognitive tasks.
\newblock {\em Nature Communications}, 13(1):90, January 2022.
\newblock Number: 1 Publisher: Nature Publishing Group.

\bibitem{garcia-cabezas_mirror_2017}
Miguel~A. García-Cabezas, Mary Kate~P. Joyce, Yohan~J. John, Basilis
  Zikopoulos, and Helen Barbas.
\newblock Mirror trends of plasticity and stability indicators in primate
  prefrontal cortex.
\newblock {\em European Journal of Neuroscience}, 46(8):2392--2405, 2017.
\newblock \_eprint: https://onlinelibrary.wiley.com/doi/pdf/10.1111/ejn.13706.

\bibitem{duncan_adaptive_2001}
John Duncan.
\newblock An adaptive coding model of neural function in prefrontal cortex.
\newblock {\em Nature Reviews Neuroscience}, 2(11):820--829, November 2001.
\newblock Number: 11 Publisher: Nature Publishing Group.

\bibitem{crowe_rapid_2010}
David~A. Crowe, Bruno~B. Averbeck, and Matthew~V. Chafee.
\newblock Rapid {Sequences} of {Population} {Activity} {Patterns} {Dynamically}
  {Encode} {Task}-{Critical} {Spatial} {Information} in {Parietal} {Cortex}.
\newblock {\em Journal of Neuroscience}, 30(35):11640--11653, September 2010.
\newblock Publisher: Society for Neuroscience Section: Articles.

\bibitem{meyers_dynamic_2008}
Ethan~M. Meyers, David~J. Freedman, Gabriel Kreiman, Earl~K. Miller, and Tomaso
  Poggio.
\newblock Dynamic {Population} {Coding} of {Category} {Information} in
  {Inferior} {Temporal} and {Prefrontal} {Cortex}.
\newblock {\em Journal of Neurophysiology}, 100(3):1407--1419, September 2008.
\newblock Publisher: American Physiological Society.

\bibitem{achterberg_one-shot_2022}
Jascha Achterberg, Mikiko Kadohisa, Kei Watanabe, Makoto Kusunoki, Mark~J.
  Buckley, and John Duncan.
\newblock A {One}-{Shot} {Shift} from {Explore} to {Exploit} in {Monkey}
  {Prefrontal} {Cortex}.
\newblock {\em Journal of Neuroscience}, 42(2):276--287, January 2022.
\newblock Publisher: Society for Neuroscience Section: Research Articles.

\bibitem{duncan_multiple-demand_2010}
John Duncan.
\newblock The multiple-demand ({MD}) system of the primate brain: mental
  programs for intelligent behaviour.
\newblock {\em Trends in Cognitive Sciences}, 14(4):172--179, April 2010.

\bibitem{sakagami_encoding_1994}
Masamichi Sakagami and Hiroaki Niki.
\newblock Encoding of behavioral significance of visual stimuli by primate
  prefrontal neurons: relation to relevant task conditions.
\newblock {\em Experimental Brain Research}, 97(3):423--436, January 1994.

\bibitem{rainer_selective_1998}
Gregor Rainer, Wael~F. Asaad, and Earl~K. Miller.
\newblock Selective representation of relevant information by neurons in the
  primate prefrontal cortex.
\newblock {\em Nature}, 393(6685):577--579, June 1998.
\newblock Number: 6685 Publisher: Nature Publishing Group.

\bibitem{buschman_synchronous_2012}
Timothy~J. Buschman, Eric~L. Denovellis, Cinira Diogo, Daniel Bullock, and
  Earl~K. Miller.
\newblock Synchronous {Oscillatory} {Neural} {Ensembles} for {Rules} in the
  {Prefrontal} {Cortex}.
\newblock {\em Neuron}, 76(4):838--846, November 2012.

\bibitem{lindsay_attention_2020}
Grace~W. Lindsay.
\newblock Attention in {Psychology}, {Neuroscience}, and {Machine} {Learning}.
\newblock {\em Frontiers in Computational Neuroscience}, 14, 2020.

\bibitem{wallis_single_2001}
Jonathan~D. Wallis, Kathleen~C. Anderson, and Earl~K. Miller.
\newblock Single neurons in prefrontal cortex encode abstract rules.
\newblock {\em Nature}, 411(6840):953--956, June 2001.
\newblock Number: 6840 Publisher: Nature Publishing Group.

\bibitem{smolensky_tensor_1990}
Paul Smolensky.
\newblock Tensor product variable binding and the representation of symbolic
  structures in connectionist systems.
\newblock {\em Artificial Intelligence}, 46(1):159--216, November 1990.

\bibitem{botvinick_reinforcement_2019}
Matthew Botvinick, Sam Ritter, Jane~X. Wang, Zeb Kurth-Nelson, Charles
  Blundell, and Demis Hassabis.
\newblock Reinforcement {Learning}, {Fast} and {Slow}.
\newblock {\em Trends in Cognitive Sciences}, 23(5):408--422, May 2019.
\newblock Publisher: Elsevier.

\bibitem{von_oswald_transformers_2022}
Johannes von Oswald, Eyvind Niklasson, Ettore Randazzo, João Sacramento,
  Alexander Mordvintsev, Andrey Zhmoginov, and Max Vladymyrov.
\newblock Transformers learn in-context by gradient descent, December 2022.
\newblock arXiv:2212.07677 [cs].

\bibitem{wang_prefrontal_2018}
Jane~X. Wang, Zeb Kurth-Nelson, Dharshan Kumaran, Dhruva Tirumala, Hubert
  Soyer, Joel~Z. Leibo, Demis Hassabis, and Matthew Botvinick.
\newblock Prefrontal cortex as a meta-reinforcement learning system.
\newblock {\em Nature Neuroscience}, 21(6):860--868, June 2018.
\newblock Number: 6 Publisher: Nature Publishing Group.

\bibitem{vyas_computation_2020}
Saurabh Vyas, Matthew~D. Golub, David Sussillo, and Krishna~V. Shenoy.
\newblock Computation {Through} {Neural} {Population} {Dynamics}.
\newblock {\em Annual Review of Neuroscience}, 43(1):249--275, 2020.
\newblock \_eprint: https://doi.org/10.1146/annurev-neuro-092619-094115.

\bibitem{buschman_goal-direction_2014}
Timothy~J. Buschman and Earl~K. Miller.
\newblock Goal-direction and top-down control.
\newblock {\em Philosophical Transactions of the Royal Society B: Biological
  Sciences}, 369(1655):20130471, November 2014.
\newblock Publisher: Royal Society.

\bibitem{macdowell_multiplexed_2023}
Camden~J. MacDowell, Alexandra Libby, Caroline~I. Jahn, Sina Tafazoli, and
  Timothy~J. Buschman.
\newblock Multiplexed {Subspaces} {Route} {Neural} {Activity} {Across}
  {Brain}-wide {Networks}, February 2023.
\newblock bioRxiv: 2023.02.08.527772 Section: New Results.

\bibitem{whittington_tolman-eichenbaum_2020}
James C.~R. Whittington, Timothy~H. Muller, Shirley Mark, Guifen Chen, Caswell
  Barry, Neil Burgess, and Timothy E.~J. Behrens.
\newblock The {Tolman}-{Eichenbaum} {Machine}: {Unifying} {Space} and
  {Relational} {Memory} through {Generalization} in the {Hippocampal}
  {Formation}.
\newblock {\em Cell}, 183(5):1249--1263.e23, November 2020.

\bibitem{dekker_determinants_2022}
Ronald~Boris Dekker, Fabian Otto, and Christopher Summerfield.
\newblock Determinants of human compositional generalization, March 2022.

\bibitem{whittington_relating_2022}
James C.~R. Whittington, Joseph Warren, and Timothy E.~J. Behrens.
\newblock Relating transformers to models and neural representations of the
  hippocampal formation, March 2022.
\newblock arXiv:2112.04035 [cs, q-bio].

\bibitem{masse_circuit_2019}
Nicolas~Y. Masse, Guangyu~R. Yang, H.~Francis Song, Xiao-Jing Wang, and
  David~J. Freedman.
\newblock Circuit mechanisms for the maintenance and manipulation of
  information in working memory.
\newblock {\em Nature Neuroscience}, 22(7):1159--1167, July 2019.
\newblock Number: 7 Publisher: Nature Publishing Group.

\bibitem{falandays_potential_2023}
J.~Benjamin Falandays, Jeff Yoshimi, William Warren, and Michael Spivey.
\newblock A {Potential} {Mechanism} for {Gibsonian} {Resonance}: {Behavioral}
  {Entrainment} {Emerges} from {Local} {Homeostasis} in an {Unsupervised}
  {Reservoir} {Network}, February 2023.

\bibitem{nelli_neural_2023}
Stephanie Nelli, Lukas Braun, Tsvetomira Dumbalska, Andrew Saxe, and
  Christopher Summerfield.
\newblock Neural knowledge assembly in humans and neural networks.
\newblock {\em Neuron}, 0(0), March 2023.
\newblock Publisher: Elsevier.

\bibitem{lowe_putting_2020}
Sindy Löwe, Peter O'Connor, and Bastiaan~S. Veeling.
\newblock Putting {An} {End} to {End}-to-{End}: {Gradient}-{Isolated}
  {Learning} of {Representations}, January 2020.
\newblock arXiv:1905.11786 [cs, stat].

\bibitem{ren_scaling_2023}
Mengye Ren, Simon Kornblith, Renjie Liao, and Geoffrey Hinton.
\newblock Scaling {Forward} {Gradient} {With} {Local} {Losses}, March 2023.
\newblock arXiv:2210.03310 [cs].

\bibitem{hinton_forward-forward_2022}
Geoffrey Hinton.
\newblock The {Forward}-{Forward} {Algorithm}: {Some} {Preliminary}
  {Investigations}, December 2022.
\newblock arXiv:2212.13345 [cs].

\bibitem{hassabis_neuroscience-inspired_2017}
Demis Hassabis, Dharshan Kumaran, Christopher Summerfield, and Matthew
  Botvinick.
\newblock Neuroscience-{Inspired} {Artificial} {Intelligence}.
\newblock {\em Neuron}, 95(2):245--258, July 2017.

\bibitem{zador_toward_2023}
Anthony Zador, Sean Escola, Blake Richards, Bence Ölveczky, Yoshua Bengio,
  Kwabena Boahen, Matthew Botvinick, Dmitri Chklovskii, Anne Churchland,
  Claudia Clopath, James DiCarlo, Surya Ganguli, Jeff Hawkins, Konrad Koerding,
  Alexei Koulakov, Yann LeCun, Timothy Lillicrap, Adam Marblestone, Bruno
  Olshausen, Alexandre Pouget, Cristina Savin, Terrence Sejnowski, Eero
  Simoncelli, Sara Solla, David Sussillo, Andreas~S. Tolias, and Doris Tsao.
\newblock Toward {Next}-{Generation} {Artificial} {Intelligence}: {Catalyzing}
  the {NeuroAI} {Revolution}, February 2023.
\newblock arXiv:2210.08340 [cs, q-bio].

\end{thebibliography}

\end{document}